# Frictional Authors

Devlin Gualtieri
Tikalon LLC, Ledgewood, New Jersey
*(gualtieri@ieee.org)*

*I present a method for text analysis based on an analogy with the dynamic friction of sliding surfaces. One surface is an array of points with a 'friction coefficient' derived from the distribution frequency of a text's alphabetic characters. The other surface is a test patch having points with this friction coefficient equal to a median value. Examples are presented from an analysis of a broad range of public domain texts, and comparison is made to the Flesch Reading Ease. Source code for the analysis program is provided.*

## Introduction

Statistical analysis of texts has a long history. Perhaps the earliest example is the 1887 paper by Mendenhall, who researched the concept of word spectrum; that is, the frequency distribution of word length in a text.[1] There's also research related to the century-old conjecture that the Gospel of John in the Bible was authored by more than one individual.[2] The question of authorship of Shakespeare's work is a popular topic. This is illustrated in a paper by Ronald Thisted and Bradley Efron in their analysis of a previously unknown nine-stanza poem attributed to Shakespeare.[3] In 1963, Mosteller and Wallace did an analysis of the Federalist Papers, a series of publications authored anonymously under the pseudonym, Publius, that identified their authors as either Alexander Hamilton, John Jay, or James Madison.[4]

A 1939 study by G. Udny Yule used sentence length as a metric,[5] Yule's paper was followed closely by a 1940 paper by C. B. Williams that looked more deeply into the statistics of this same metric.[6] In a 1975 paper, H. S. Sichel further examined the distribution of word frequencies,[7] and Claude S. Brinegar investigated the possible authorship of 10 letters, published in 1861, and attributed to Mark Twain. Brinegar used word-length frequency to confirm Twain's authorship.[8] One of the most popular text analysis tools is the Flesch Reading Ease formula that scores a text's readability on a scale of 0-100.[9] The Flesch–Kincaid grade level gives a US school grade level needed to read a text based on total words, total sentences, and total syllables.[10]

## Letter Frequency

There's an uneven distribution of the alphabetic characters, *a-z*, of the English language, with some characters, such as *e* and *t*, used more often in words than others, such as *j*, *q*, and *x*.[11][12] This frequency distribution is summarized in Table I.

Table I. Frequency of use for alphabet letters for English.[13] The scaled complement inverts the distribution (see text).

| Letter | Frequency | Scaled Complement | Letter | Frequency | Scaled Complement |
|---|---|---|---|---|---|
| A | 0.0850 | 0.2427 | N | 0.0665 | 0.4115 |
| B | 0.0207 | 0.8294 | O | 0.0716 | 0.3650 |
| C | 0.0454 | 0.6040 | P | 0.0317 | 0.7290 |
| D | 0.0338 | 0.7099 | Q | 0.0020 | 1.0000 |
| E | 0.1116 | 0.0000 | R | 0.0758 | 0.3266 |
| F | 0.0181 | 0.8531 | S | 0.0574 | 0.4945 |
| G | 0.0247 | 0.7929 | T | 0.0695 | 0.3841 |
| H | 0.0300 | 0.7445 | U | 0.0363 | 0.6870 |
| I | 0.0754 | 0.3303 | V | 0.0101 | 0.9261 |
| J | 0.0020 | 1.0000 | W | 0.0129 | 0.9005 |
| K | 0.0110 | 0.9179 | X | 0.0029 | 0.9918 |
| L | 0.0549 | 0.5173 | Y | 0.0178 | 0.8558 |
| M | 0.0301 | 0.7436 | Z | 0.0027 | 0.9936 |

## Text Sources

Validation of text analysis models is aided by the large number and variety of open source material on the Internet. One such source is Project Gutenberg, started in 1971 by Michael S. Hart.[14] The most current published statistic of July 20, 2011, lists 36,701 books in the US Project Gutenberg archive.[15] Table II lists the texts used in this study. Also listed are the Flesch Reading Ease score and the equivalent US school grade level.

Table II. Texts used in this study, along with their Flesch Reading Ease score and the equivalent US school grade level.

| Title | Author | Ease | Grade |
|---|---|---|---|
| Aeneid[16] | Virgil | 69.65 | 10.2 |
| Alice's Adventures in Wonderland[17] | Lewis Carrol | 87.55 | 5.4 |
| The Hound of the Baskervilles[18] | Arthur Conan Doyle | 80.92 | 5.9 |
| Beowulf[19] | Beowulf Poet | 67.38 | 9.0 |
| The Call of the Wild[20] | Jack London | 77.37 | 7.2 |
| A Christmas Carol[21] | Charles Dickens | 80.01 | 6.2 |
| David Copperfield[22] | Charles Dickens | 78.69 | 6.7 |
| Robinson Crusoe[23] | Daniel Defoe | 54.16 | 18.2 |
| The Federalist Papers[24] | Publius (Alexander Hamilton, James Madison, and John Jay) | 38.39 | 16.0 |
| Frankenstein[25] | Mary Shelley | 64.85 | 10.0 |
| Great Expectations[26] | Charles Dickens | 77.16 | 7.3 |
| Huckleberry Finn[27] | Mark Twain | 85.62 | 6.1 |
| Iliad[28] | Homer | 66.85 | 13.4 |
| A Portrait of the Artist as a Young Man[29] | James Joyce | 79.40 | 6.5 |
| Leviathan[30] | Thomas Hobbes | 49.32 | 15.9 |
| The Theory of The Leisure Class[31] | Thorstein Veblen | 37.47 | 16.4 |
| Moby Dick[32] | Herman Melville | 72.90 | 9.0 |
| Oliver Twist[33] | Charles Dickens | 78.69 | 6.7 |
| The Wonderful Wizard of Oz[34] | L. Frank Baum | 86.13 | 5.9 |
| Paradise Lost[35] | John Milton | 45.80 | 19.4 |
| The Natural History, Volume VI[36] | Pliny the Elder | 59.16 | 12.2 |
| Winnie-the-Pooh[37] | A. A. Milne | 92.42 | 3.5 |
| The Scarlet Letter: A Romance[38] | Nathaniel Hawthorne | 66.27 | 9.4 |
| Hamlet[39] | Shakespeare | 83.56 | 4.9 |
| Sonnets[40] | Shakespeare | 69.89 | 12.2 |
| Siddhartha[41] | Hermann Hesse | 75.03 | 8.1 |
| Swann's Way[42] | Marcel Proust | 59.00 | 14.3 |
| A Tale of Two Cities[43] | Charles Dickens | 78.28 | 6.9 |
| Timaeus[44] | Plato | 43.63 | 18.1 |
| The Time Machine[45] | H. G. Wells | 79.60 | 6.4 |
| Treasure Island[46] | Robert Louis Stevenson | 86.13 | 5.9 |
| Uncle Tom's Cabin[47] | Harriet Beecher Stowe | 78.18 | 6.9 |

## Sliding Friction Model

It seems likely that elementary texts, such as children's books, would be built from words having more common alphabet characters, whereas more ponderous works would have more frequent use of the less common alphabet characters.  Independent of the validity of this hypothesis, the variation of alphabetic character frequency through a text might produce a digital signature of an author, a thematic structure of a literary work, or a literature genre.

In this study, text analysis was envisioned as the sliding friction between two surfaces.  One of these is a fixed width surface having point friction coefficients relating to a scaled complement of the frequency of each letter.  The other is a square patch of the same width for which these friction coefficients are the same and set to a median value.  The texts were pre-processed to convert all alphabet characters to lower case, and to remove all non-alphabet characters; e. g., spaces, punctuation and line-feeds.  This sliding friction model is shown schematically in fig. 1.

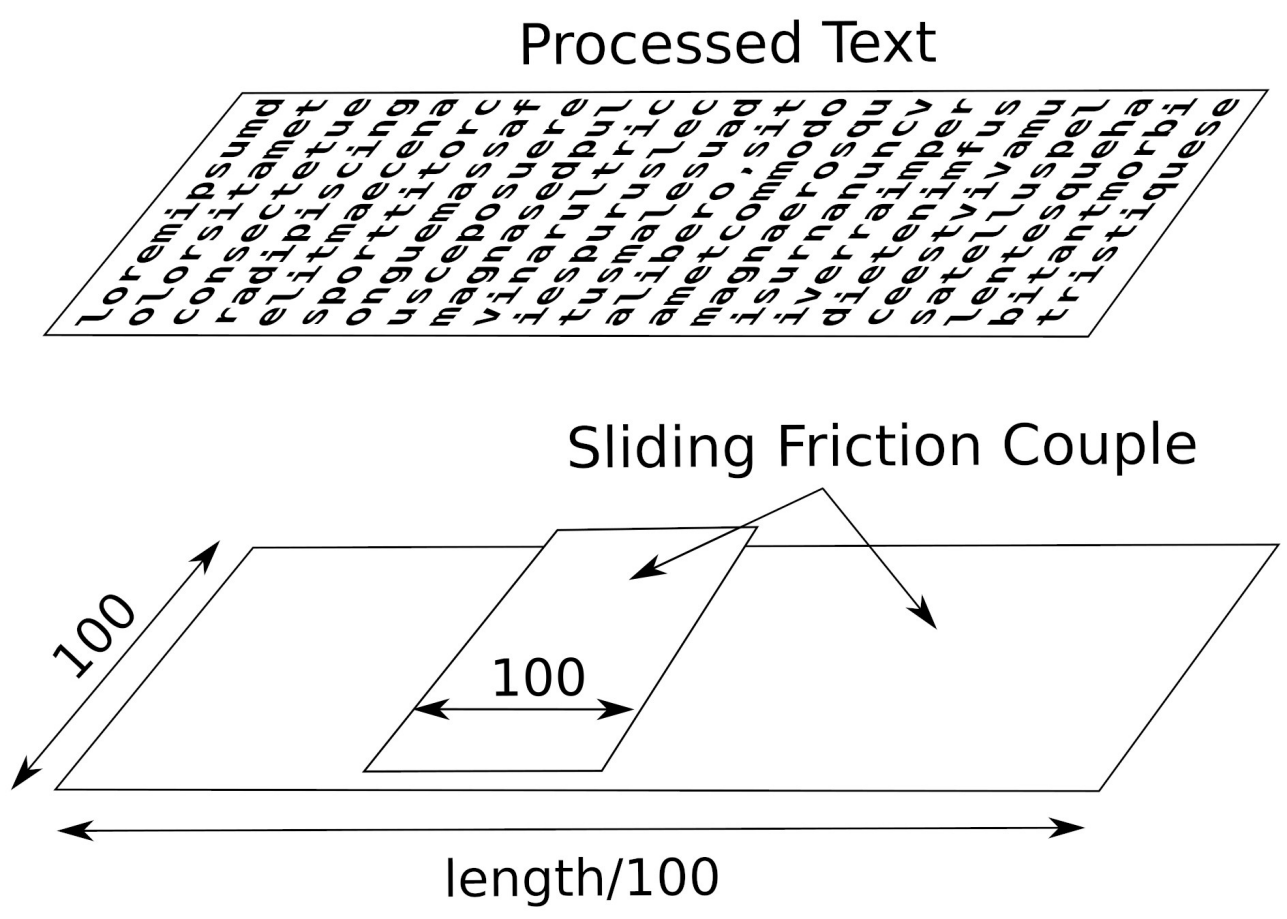

Fig. 1.  Sliding friction model.

The text is  pre-processed to convert all alphabet characters to lower case, and all non-alphabet characters are removed.  The text is then mapped to a surface of width 100, and the alphabet characters are converted to a friction coefficient, as described in the text.

Another surface of the same width and a uniform friction coefficient is moved along the text surface to generate the sliding friction values of the text.  See the source code in Appendix II for details.

The friction coefficient of each alphabet character was derived as a complement of the letter frequency for which more frequent characters had low friction, while less frequent characters had high friction. If **f** denotes the letter frequency in Table I, the scaled complement **SC** is given as

$$SC = ((1-f)-min)/(max-min) \quad (1)$$

in which **min** is the minimum value of (1-**f**), 0.8884, and **max** is the maximum value of (1-f), 0.9980. This scales **SC** from 0 to 1, as shown in Table I. This gives the most frequent character, *e*, a scaled complement of 0, and the least frequent character, *q*, a scaled complement of 1, and these scaled complements are used as the friction coefficients for each letter.

Text Analysis

Two analysis programs were developed, and their source code can be found in the appendices. A Python program was used to determine the Flesch Reading Ease and Grade Level. This program was enabled by the *textstat* library written by Shivam Bansal and Chaitanya Aggarwal[48] An analysis program was developed in the C programming language to convert ASCII text files to friction coefficients, perform the sliding friction function, and write the data to files. Note that the text files on Project Gutenberg are encoded as UTF-8, and they must be converted to ASCII. This can be done using the Linux command line

```
uni2ascii -B utf-8.txt >asc.txt
```
(2)

in which *utf-8.txt* is the input UTF-8 file, and *asc.txt* is the output ASCII file.

Fig. 2 shows two examples of the variation in friction over the length of the text. These trends, illustrated by other examples in a later section, are clearly distinguished from noise and indicate structure. Fig. 3 shows the correlation between the text mean friction **MF** of the studied works and the Flesch Reading Ease. The line is the linear regression of these data, given as

$$Ease = -687 + (0.225\ MF) \quad (3)$$

The average standard deviation of the mean friction values for these texts is 23.6. A histogram of the standard deviations is shown as fig. 4. Fig. 5 is a summary comparison of the measured and predicted reading ease.

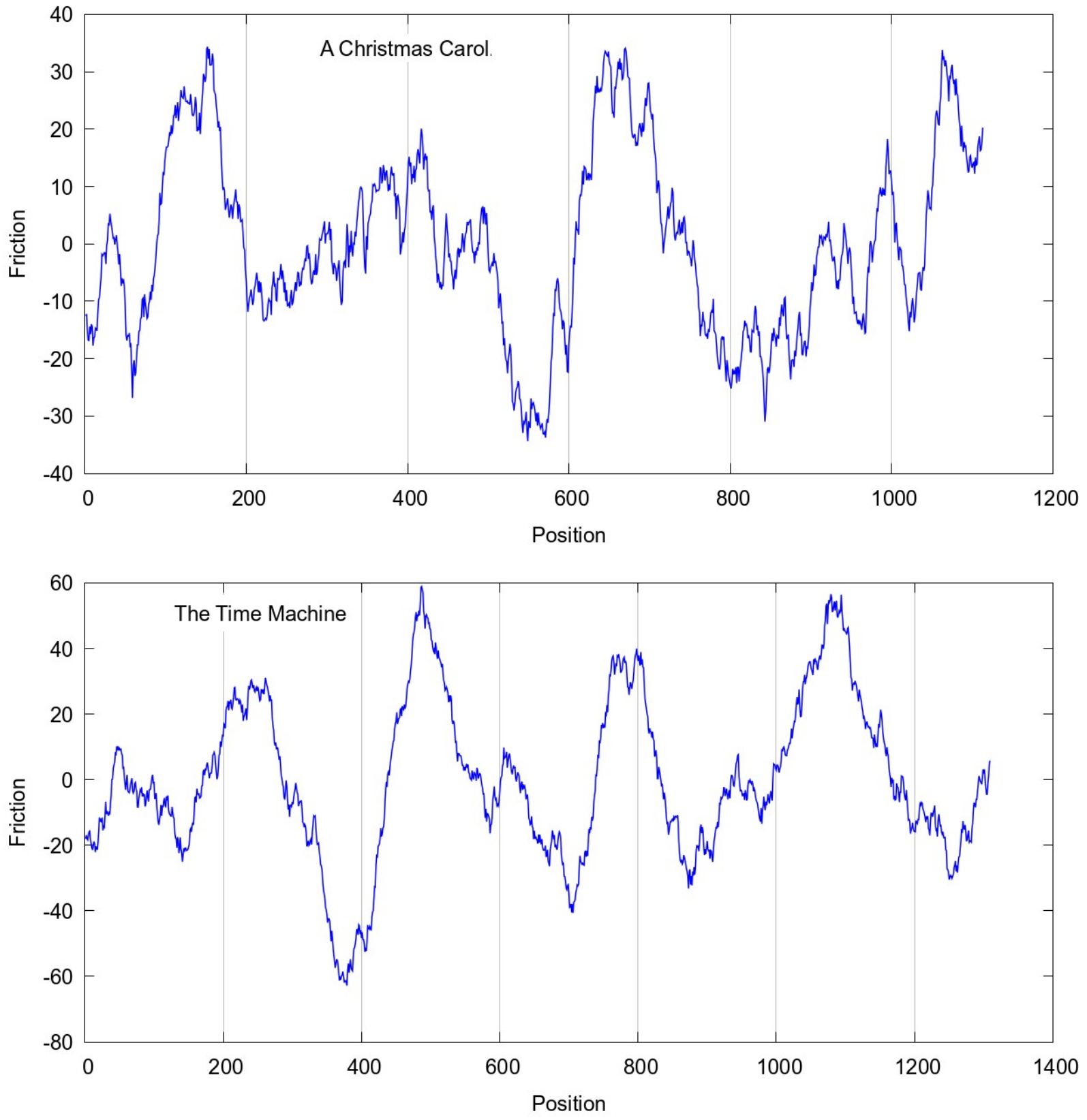

Fig. 2. Text friction across the length of Charles Dickens', "A Christmas Carol" (top) and H. G. Wells', "The Time Machine" (bottom).

The friction appears to distinguish different sections of the narratives.

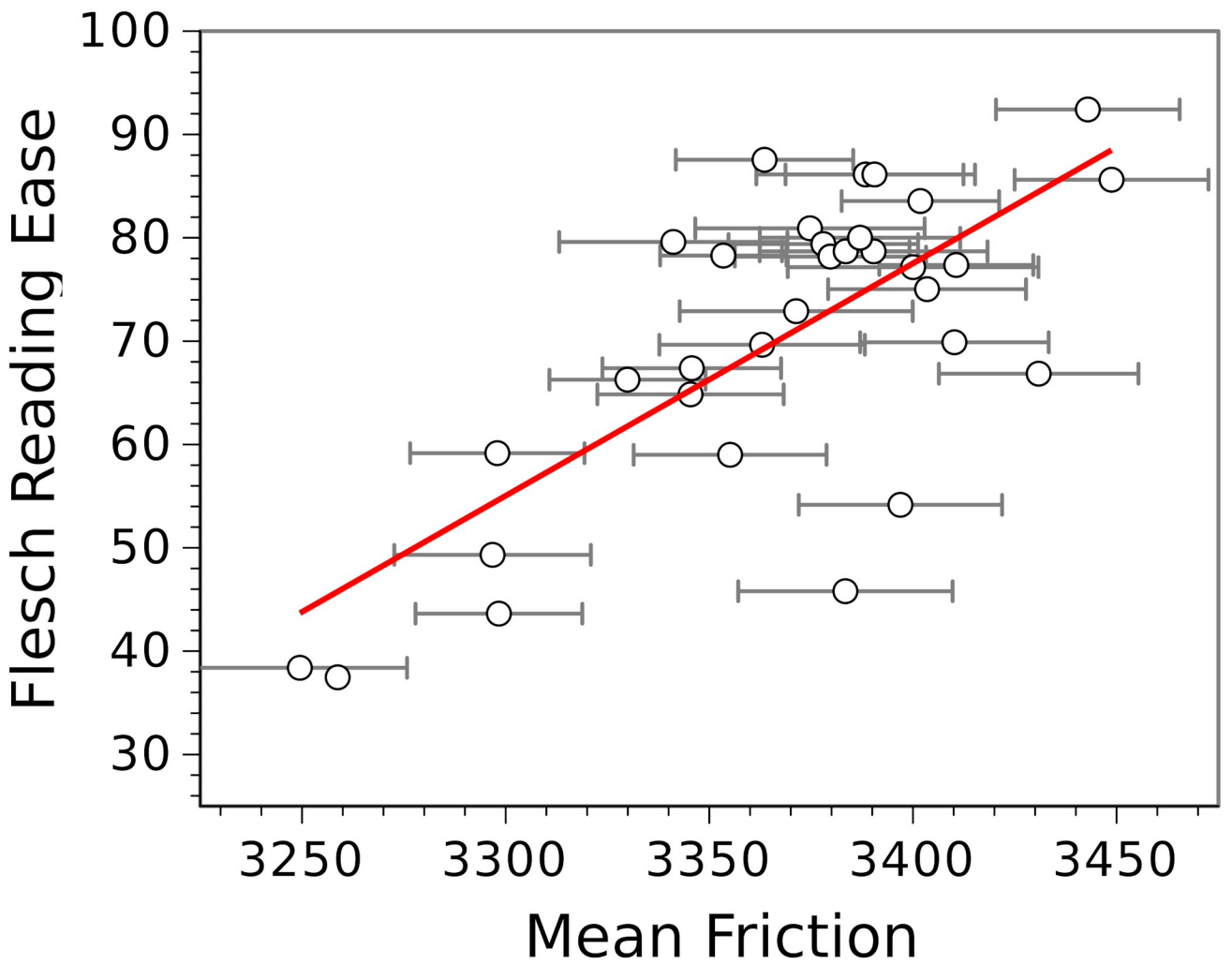

Fig. 3. Flesch Reading Ease as a function of the mean text friction. The linear regression is shown by the line. The standard deviation of friction in each text are shown.

The smallest standard deviation was for Shakespeare's Hamlet. The largest standard deviation was for Plato's Timaeus. The average standard deviation of the mean friction values for these texts is 23.6.

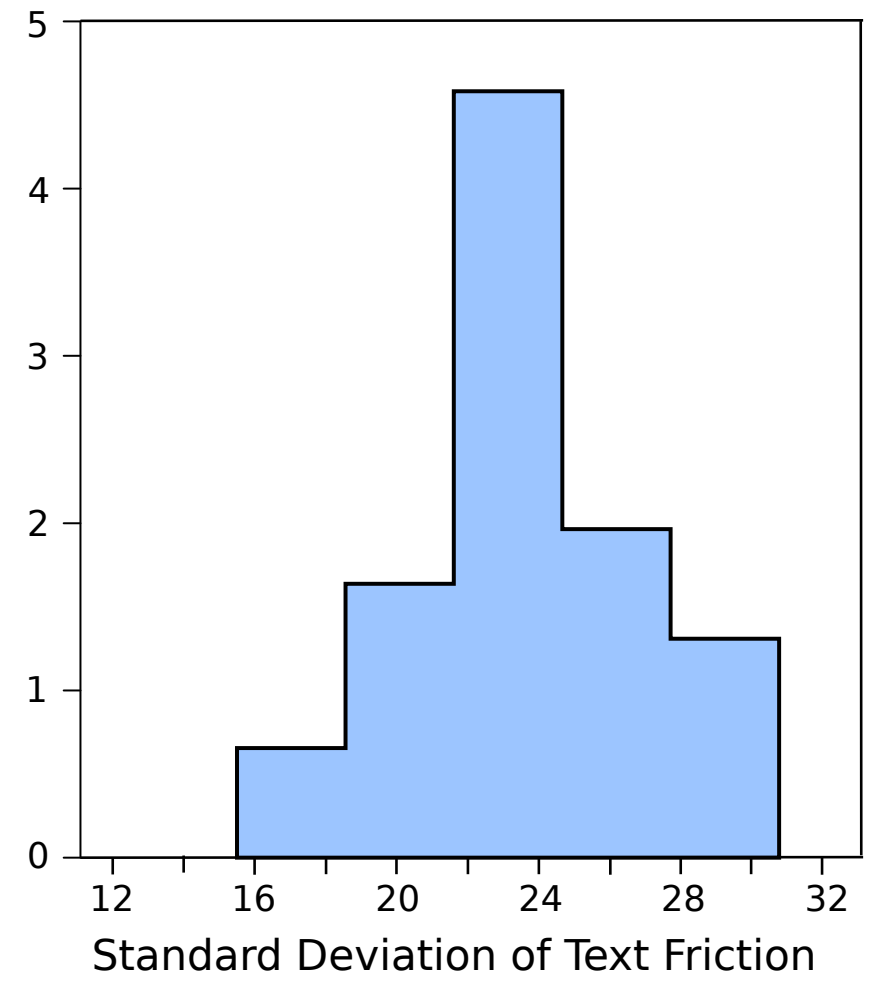

Fig. 4. The histogram of the standard deviation of the friction values of each text.

The smallest standard deviation was for Shakespeare's Hamlet (15.50). The largest standard deviation was for Plato's Timaeus (28.61).

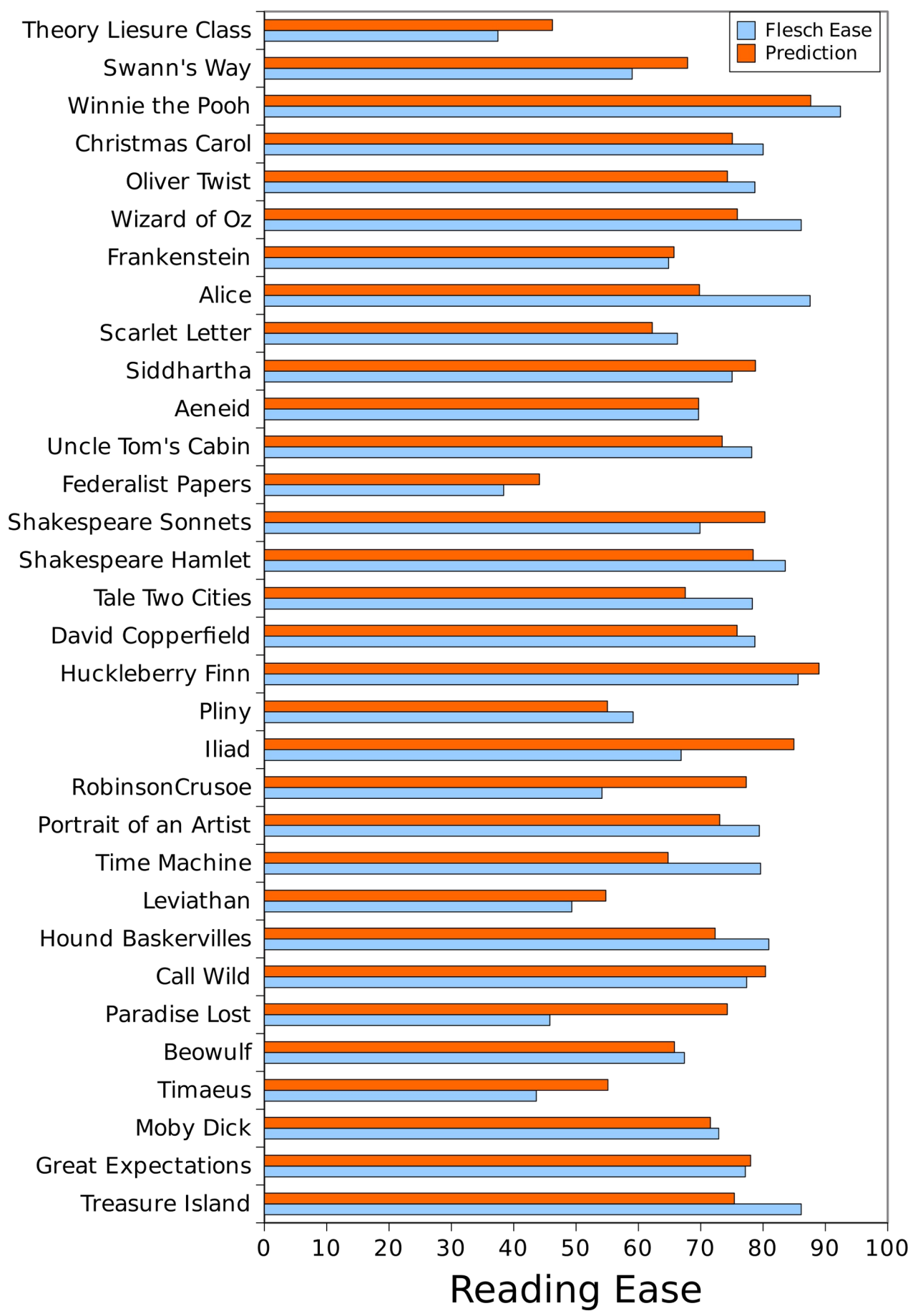

Fig. 5. Predicted and measured reading ease for the 32 texts used in this study.

The predicted reading ease was calculated using the linear regression of eq. 3.

## Compendium of Results

Fig. 6 shows the variation of friction through each of the studied texts.

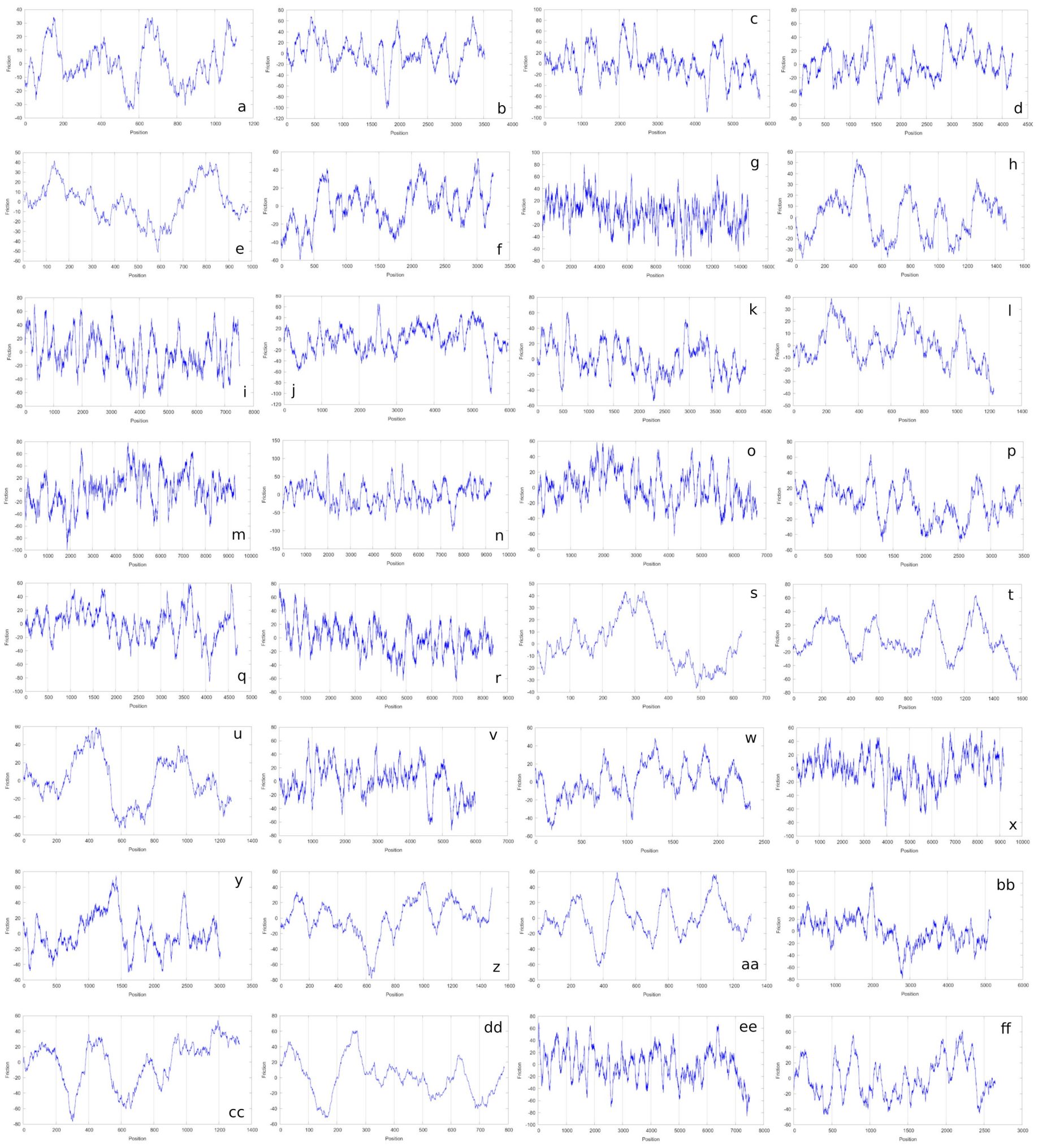

Fig. 6. A compendium of the friction values along the lengths of the 32 studied texts. (a) A Christmas Carol, (b) A Portrait of the Artist as a Young Man, (c) A Tale of Two Cities, (d) Aeneid, (e) Alice's Adventures in Wonderland, (f) Frankenstein, (g) David Copperfield, (h) Beowulf, (i) Great Expectations, (j) Iliad, (k) Huckleberry Finn, (l) Hamlet, (m) Leviathan, (n) Moby Dick, (o) Oliver Twist, (p) Paradise Lost, (q) Robinson Crusoe, (r) Swann's Way, (s) Shakespeare Sonnets, (t) Siddhartha, (u) The Call of the Wild, (v) The Natural History, Volume VI, (w) The Hound of the Baskervilles, (x) The Federalist Papers, (y) The Scarlet Letter: A Romance, (z) The Wonderful Wizard of Oz, (aa) The Time Machine, (bb) The Theory of The Leisure Class, (cc) Timaeus, (dd) Winnie-the-Pooh, (ee) Uncle Toms Cabin, (ff) Treasure Island

## Discussion

The concept of text friction may be a useful method of scoring readability. Its correlation with Flesch Reading Ease could be enhanced by using a non-linear scaling of character friction based on their frequencies, rather than the linear scaling used here. It's realized, however, that Flesh Reading Ease is just a conjecture, albeit a logical conjecture, so a goal of absolute correspondence is not considered to be necessary. A neural network model could be used to determine the best friction coefficient for each character.

The text friction appears to track the structure of a text, and it might be useful in distinguishing ordered narrative from more random ramblings, such as works characterized as *stream of consciousness* (compare, for example, the results for *The Time Machine* with *Swann's Way*).

One feature of plays and dialogues are the repetitive instances of character names that identify their spoken parts. This attribute dilutes the accuracy of both the Flesch analysis and this friction analysis.

<u>Appendix I</u>. Python program for batch processing of text files to produce the Flesch Reading Ease and Grade Level.

```
import sys
import os
import textstat
# textstat library by Shivam Bansal and Chaitanya Aggarwal
# see https://pypi.org/project/textstat/

#open output file in append mode
output_file = open("readability.dat", "a")

# iterate over all files
for input_file in os.listdir(os.getcwd()):
    if input_file.endswith("txt"):
        #open input file in read mode
        text_file = open(input_file, "r")

        #read whole file to a string
        text_data = text_file.read()

        #close file
        text_file.close()

        ease = textstat.flesch_reading_ease(text_data)
        grade = textstat.flesch_kincaid_grade(text_data)

        data_string = input_file +"\t"+ str(ease) + "\t" + str(grade) + "\n"

        print(data_string)

        output_file.write(data_string)
    else:
        continue

output_file.close()
```

Appendix II. Source code for the text friction analysis program.

```
/* -*- Mode: C; indent-tabs-mode: t; c-basic-offset: 4; tab-width: 4 -*-  */
/*
 * text_friction.c
 * Copyright (C) 2022 drdev <gualtieri@ieee.org>
 *
 * text_friction is free software: you can redistribute it and/or modify it
 * under the terms of the GNU General Public License as published by the
 * Free Software Foundation, either version 3 of the License, or
 * (at your option) any later version.
 *
 * text_friction is distributed in the hope that it will be useful, but
 * WITHOUT ANY WARRANTY; without even the implied warranty of
 * MERCHANTABILITY or FITNESS FOR A PARTICULAR PURPOSE.
 * See the GNU General Public License for more details.
 *
 * You should have received a copy of the GNU General Public License along
 * with this program.  If not, see <http://www.gnu.org/licenses/>.
 */

//To compile using gcc: gcc -o text_friction text_friction.c -lm

#include <stdio.h>
#include <stdlib.h>
#include <stddef.h>
#include <string.h>
#include <math.h>

int i, x, y;
int height, width;
char ch;
char *ret_value;
int pos;
char fn1[64];
char fn2[64] = "output.dat";
char fn3[64] = "statistics.dat";
FILE *indata;
FILE *outdata;
FILE *outdata2;
long int fsize;
float frict, mean;
float deviation, variance, sumsqr, stddev;
float friction_coef[26] =
    { 0.2427, 0.8294, 0.6040, 0.7099, 0.0000, 0.8531, 0.7929,
    0.7445, 0.3303, 1.0000, 0.9179, 0.5173, 0.7436, 0.4115, 0.3650, 0.7290,
      1.0000,
    0.3266, 0.4945, 0.3841, 0.6870, 0.9261, 0.9005, 0.9918, 0.8558, 0.9936
};

//median value is 0.7363

/* Prototypes */
char *strcpy(char *dest, const char *src);
char *strchr(const char *str, int c);
char *fgets(char *s, int n, FILE * f);
```

```
void exit(int status);
/* end of prototypes */

int main(int argc, char *argv[])
{

    if (argc < 2) {
      printf("Usage: text_friction input.txt\n");
      exit(1);
    }

    strcpy(fn1, argv[1]);
    printf("\nInput file selected = %s\n", fn1);

    if ((indata = fopen(fn1, "rb")) == NULL) {
      printf("\nInput file cannot be opened.\n");
      exit(1);
    }
//build filename for datafile
//somewhat brute force, but easy to understand
    strcpy(fn2, fn1);
    ret_value = strchr(fn2, '.');
    if (ret_value == NULL) {
      strcpy(fn2, "analysis.dat");
      printf("Using %s as output file\n", fn2);
    } else {
      pos = (int) (ret_value - fn2);
      fn2[pos + 1] = 'd';
      fn2[pos + 2] = 'a';
      fn2[pos + 3] = 't';
      fn2[pos + 4] = '\0';
      printf("Output file selected = %s\n", fn2);
    }

    if ((outdata = fopen(fn2, "w")) == NULL) {
      printf("\nOutput file cannot be opened.\n");
      exit(1);
    }
//open statistics file as append
    if ((outdata2 = fopen(fn3, "a")) == NULL) {
      printf("\nStatistics file cannot be opened.\n");
      exit(1);
    }
//get size of input file to allocate character array
    fseek(indata, 0L, SEEK_END);
    fsize = ftell(indata);
    rewind(indata);           //return to file start

    if (fsize != -1) {
      printf("Size of input file = %ld bytes.\n", fsize);
    }
//We use an array of pointers to create a two dimensional array to store friction
data
//our array is 100 characters wide
    width = 100;
    height = (fsize / width);

    float (*surface)[height];
```

```
    surface = malloc(sizeof(float[width][height]));

    float *friction;
    friction = malloc(height * sizeof(float));

//read character by character, process, and store
//ASCII A=65, Z=90
//ASCII a=97, z=122
    x = 0;
    y = 0;
    while ((ch = fgetc(indata)) != EOF) {
      if ((ch > 64) && (ch < 91)) {
//convert to lowercase
          ch = ch + 32;
      }
//ignore all other characters
      if ((ch > 96) && (ch < 123)) {
          printf("%c", ch);
          surface[x][y] = friction_coef[ch - 97];
          x++;
          if (x > 99) {
            x = 0;
            y++;
          }
      }
    }

//update height
    height = y - 1;

    printf("\nwidth = %d, height = %d.\n", width, height);

//do friction function sliding patch through entire array
    for (y = 0; y < (height - 100); y++) {
      frict = 0;
      for (i = 0; i < 100; i++) {
          for (x = 0; x < width; x++) {
//we slide patch with value 0.7363 through surface
            frict = frict + (surface[x][y + i] * 0.7363);
          }
      }
      friction[y] = frict;
    }

//calculate mean
    mean = 0;
    for (i = 0; i < (height - 100); i++) {
      mean = mean + friction[i];
    }
    mean = mean / (float) (height - 100);

//calculate standard deviation
    sumsqr = 0;
    for (i = 0; i < (height - 100); i++) {
      deviation = friction[i] - mean;
      sumsqr = sumsqr + (deviation * deviation);
    }
```

```
    variance = sumsqr / (float) (height - 100);
    stddev = sqrt(variance);

//write datafile
    fprintf(outdata, "%s\n", fn1);
    for (i = 0; i < (height - 100); i++) {
      fprintf(outdata, "%f\t%f\n", friction[i], friction[i] - mean);
    }

//write to statistics file
    fprintf(outdata2, "%s\t%f\t%f\n", fn1, mean, stddev);

//free memory
    free(surface);
    free(friction);

    fclose(indata);
    fclose(outdata);
    fclose(outdata2);

    printf("\n>>>Done<<<\n");

    return (0);

}
```